\documentclass[letterpaper]{article} 
\usepackage[preprint]{aaai2027}  
\usepackage[hyphens]{url}  
\usepackage{graphicx} 
\usepackage{natbib}  
\usepackage{caption} 
\usepackage{algorithm}
\usepackage{algorithmic}

\usepackage{newfloat}
\usepackage{listings}
\DeclareCaptionStyle{ruled}{labelfont=normalfont,labelsep=colon,strut=off} 
\floatstyle{ruled}
\newfloat{listing}{tb}{lst}{}
\floatname{listing}{Listing}

\usepackage{booktabs}
\usepackage{colortbl}

\usepackage{amsmath}
\usepackage{amssymb}
\usepackage[breakable,skins]{tcolorbox}
\usepackage{tikz}
\usetikzlibrary{positioning,arrows.meta,shapes.geometric,calc,decorations.pathmorphing,decorations.pathreplacing,patterns}
\DeclareMathOperator*{\argmax}{arg\,max}
\definecolor{stackrelevance}{RGB}{0,0,0}
\definecolor{stacksupport}{RGB}{0,105,100}
\definecolor{stacknote}{RGB}{92,92,92}
\definecolor{stacktablehead}{RGB}{235,238,240}
\definecolor{stacktablegroup}{RGB}{224,228,230}
\definecolor{stacktablebase}{RGB}{244,245,246}
\definecolor{stacktableours}{RGB}{250,232,225}
\newcommand{\algrel}[1]{\textcolor{stackrelevance}{#1}}
\newcommand{\algcov}[1]{\textcolor{stacksupport}{#1}}

\newcommand{\stacktablestyle}{%
  \renewcommand{\arraystretch}{1.08}%
  \arrayrulecolor{black!68}}
\newcommand{\stacktableheadrow}{\rowcolor{stacktablehead}}
\newcommand{\stacktablegrouprow}{\rowcolor{stacktablegroup}}
\newcommand{\stacktablebaserow}{\rowcolor{stacktablebase}}
\newcommand{\stacktableoursrow}{\rowcolor{stacktableours}}
\newcommand{\stacktableaccent}[1]{\textcolor{stackrelevance}{#1}}
\newcommand{\vqavtwo}{VQA\textsuperscript{v2}}
\newcommand{\vqatext}{VQA\textsuperscript{Text}}
\newtcolorbox{stackpropbox}{
  enhanced,
  breakable,
  colback=stackrelevance!4,
  colframe=stackrelevance!28,
  boxrule=.35pt,
  borderline west={1.4pt}{0pt}{stackrelevance},
  arc=1mm,
  boxsep=0pt,
  left=5pt,
  right=5pt,
  top=4pt,
  bottom=4pt,
  before skip=5pt,
  after skip=5pt
}
\newtcolorbox{stacktheorembox}{
  enhanced,
  breakable,
  colback=stacksupport!4,
  colframe=stacksupport!28,
  boxrule=.35pt,
  borderline west={1.4pt}{0pt}{stacksupport},
  arc=1mm,
  boxsep=0pt,
  left=5pt,
  right=5pt,
  top=4pt,
  bottom=4pt,
  before skip=5pt,
  after skip=5pt
}

\title{StackTok: Accelerating VLMs Inference with Budget-Adaptive \\Visual Token Selection}
\author{\fontsize{12}{14}\selectfont\mbox{Zhenbin Wang, Lei Zhang$^{*}$, Lituan Wang, Wei Huang, Yan Wang, Zhenwei Zhang}}
\affiliations{Sichuan University\\\texttt{wangzhenbin@stu.scu.edu.cn}}

\newcommand{\stackcodeurl}[1]{%
\leavevmode\pdfstartlink attr{/Border [0 0 0]} user{/Subtype /Link /A << /Type /Action /S /URI /URI (\pdfescapestring{#1}) >>}%
\url{#1}\pdfendlink}
\begin{document}

\maketitle
\begingroup
\renewcommand{\thefootnote}{\fnsymbol{footnote}}
\footnotetext[1]{The corresponding author\\
\hspace*{1.8em}Code: \stackcodeurl{https://github.com/wongzbb/StackToK}}
\endgroup

\begin{abstract}
 Increasing image resolution produces ever-longer visual-token sequences in vision-language models (VLMs), substantially raising their inference cost. To reduce this overhead without retraining, existing methods select compact token subsets that prioritize query relevance, visual coverage, or a fixed trade-off between them. The appropriate balance, however, varies across queries and token budgets: localized questions favor relevance, whereas holistic questions demand broader visual coverage. We introduce StackTok, a training-free selector that treats query relevance as the objective and visual coverage as budget-calibrated support. StackTok builds a size-indexed coverage reference from a coverage-only greedy sequence and adjusts its support target using query--vision affinity entropy. A reference-gated interleaved selection policy then switches between relevance- and coverage-oriented additions according to the current subset's support deficit. For high-resolution inputs, StackTok allocates one shared token budget across crops according to the combined marginal gain of locally nominated tokens. Evaluated with five VLMs over ten distinct image-understanding benchmarks, StackTok ranks first among training-free selectors in every tested model--budget setting. On high-resolution LLaVA-NeXT-7B, it retains 95.26\% of full-token performance with only 160 of 2{,}880 (5.6\%) visual tokens.
\end{abstract}

\section{Introduction}

Recent VLMs improve fine-grained visual understanding by encoding images at increasingly high resolutions. A $336\times336$ image produces 576 visual tokens in LLaVA-1.5~\citep{liu2023llava}, while the multi-crop encoder of LLaVA-NeXT~\citep{liu2024llavanext} can produce up to 2{,}880. These tokens enter the language backbone together with the prompt, so long visual sequences substantially increase prefilling latency and memory consumption. Training-free token selection offers a direct remedy by retaining a compact subset before large language model (LLM) inference without adding parameters or fine-tuning~\citep{chen2024fastv,yang2025visionzip}. This practical bottleneck raises a more fundamental question: when only a limited token budget can be retained, what makes a subset of visual tokens informative?

Existing selectors provide two complementary answers. Relevance-oriented methods use language-model attention or instruction tokens to preserve evidence related to the query~\citep{chen2024fastv,zhang2024sparsevlm}. They can retain fine details from a queried region, but may discard contextual evidence elsewhere in the image. Coverage-oriented methods instead preserve representative or diverse tokens to retain a broader view of the scene~\citep{yang2025visionzip,alvar2025divprune}. They support holistic understanding, but may spend a tight budget on regions unrelated to the question. Reading a sign and describing an entire scene therefore call for different subsets. MMTok~\citep{dong2025mmtok} takes an important step by combining query relevance and visual coverage in a unified submodular objective. Its fixed coefficient, however, assigns the two signals the same exchange rate across queries, retained-set sizes, and selection states. Fixed scalarization thus leaves unresolved how much visual coverage is appropriate for the current input and subset size.

\begin{figure*}[t]
\centering
\includegraphics[width=0.99\textwidth]{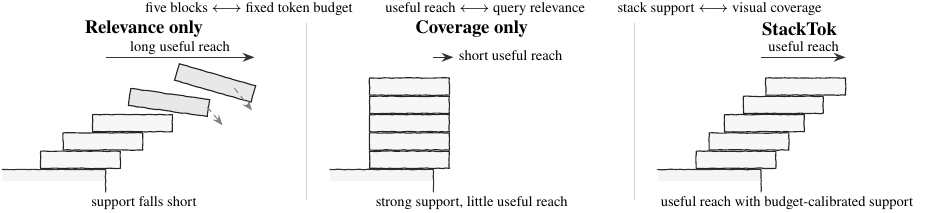}

\caption{\textbf{Block-stacking intuition for visual-token selection.}
All three configurations use the same number of blocks, corresponding to the same fixed token budget. Relevance-only placement pursues a long useful reach but can leave the stack without enough support. Coverage-only placement preserves strong support but makes little progress toward the target. StackTok pursues useful reach while using a budget-calibrated support requirement to determine when visual coverage should be prioritized. Useful reach and stack support correspond to query relevance and visual coverage, respectively.}
\label{fig:motivation}
\end{figure*}

The deeper issue is that query relevance and visual coverage do not play symmetric roles. Figure~\ref{fig:motivation} illustrates this asymmetry through the maximum-overhang problem in block stacking~\citep{paterson2009overhang}. Given a fixed number of blocks, the goal is to extend the stack toward a target without giving up the support needed to keep it standing. Relevance-only placement pushes every block outward and achieves a long reach, but can leave the stack poorly supported. Coverage-only placement keeps the blocks closely aligned and the stack well supported, but makes little useful progress. StackTok follows a different principle: pursue useful reach while treating support as a requirement calibrated to the number of blocks already placed. In visual-token selection, useful reach corresponds to query relevance, support corresponds to visual coverage, and the number of blocks corresponds to the retained-token budget. This analogy suggests maximizing relevance while treating coverage as a budget-calibrated support requirement, rather than assigning the two signals a fixed exchange rate.

Guided by this view, we introduce StackTok, a training-free selector that uses the visual coverage of the current subset to determine which signal guides the next selection. StackTok first builds a size-indexed coverage reference curve from a coverage-only greedy sequence. It then uses the normalized entropy of query--vision affinities to adjust the support target for each input. At each step, StackTok prioritizes query relevance when the current subset reaches this target, and visual coverage otherwise. Reassessing this state-dependent decision after every addition yields a single interleaved sequence whose priorities evolve with the input and selected subset. For high-resolution inputs, StackTok further allocates a shared token budget across crops according to their combined marginal gains. Our contributions are threefold:
\begin{itemize}
\item We recast training-free visual-token selection as an asymmetric objective-and-support problem. Query relevance defines the objective, while a subset-size-indexed coverage reference provides the visual support criterion instead of a fixed exchange rate between the two signals.
\item We introduce StackTok, a training-free selector that builds this reference from a coverage-only greedy sequence, calibrates its strictness from query--vision affinity entropy, and uses the current subset to interleave relevance- and coverage-oriented additions under fixed retention budget.
\item We extend StackTok to shared-budget multi-crop selection. Across five VLMs, 
ten distinct benchmarks, StackTok ranks first among training-free selectors in every tested model--budget setting.
\end{itemize}

\begin{figure*}[t]
\centering
\includegraphics[width=\textwidth]{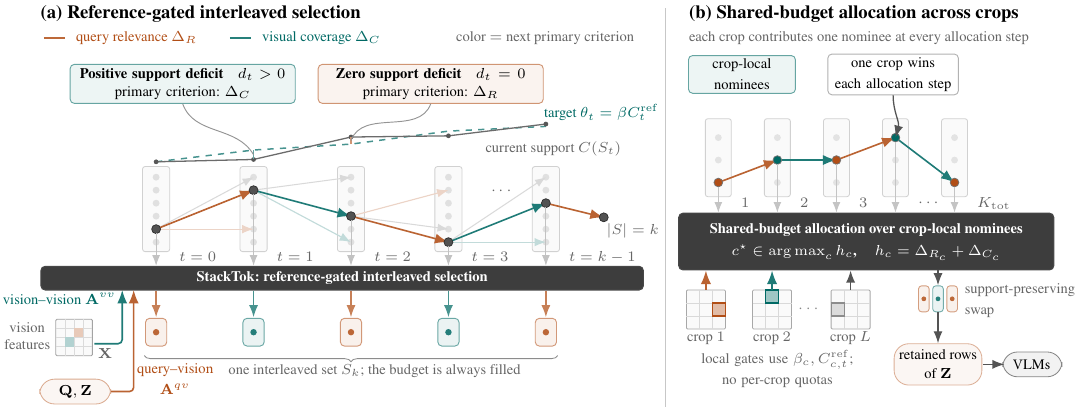}
\vspace{-1.0em}

\caption{\textbf{Overview of StackTok.}
(a) Each column depicts the candidates available at a retained-set size. The support deficit selects a coverage-oriented ($\Delta_C$) or relevance-oriented ($\Delta_R$) primary criterion for the next addition; teal and orange encode that criterion. The size-indexed target gates the primary criterion rather than constraining every prefix, and the coverage-only greedy sequence is used only to construct the reference curve. (b) For high-resolution inputs, crop-local gates nominate candidates and shared-budget allocation compares only those nominees. Faded dots and branches denote unchosen candidates. After the fixed budget is filled, optional support-preserving swap refinement and restoration of the original spatial order send the retained rows of $\mathbf Z$ to the VLMs for inference.}
\vspace{-0.8em}
\label{fig:framework}
\end{figure*}

\section{Related Work}

\subsection{Efficient Inference for Large Language and Multimodal Models}
Efficient LLM inference has been advanced by input/output (I/O)-aware attention kernels, memory-aware serving systems, and context-management methods, which reduce computation or memory costs within the language backbone~\citep{dao2022flashattention,kwon2023vllm,xiao2023streamingllm}. VLMs introduce a complementary bottleneck by coupling this backbone with non-textual inputs~\citep{mllm_gpt4v,team2023gemini}. High-resolution, multi-crop, and video inputs can produce hundreds or thousands of visual tokens that are processed together with text~\citep{liu2023llava,liu2024llavanext,bai2025qwenvl,lin2024videollava}. Learned resampling and pooling modules reduce this visual input, but typically require training or architectural modification~\citep{alayrac2022flamingo,li2024llamavid,yao2024deco}. Training-free visual-token reduction instead shortens the input to a pretrained language backbone and is orthogonal to these model-level and system-level optimizations.

\subsection{Visual Token Reduction for Vision-Language Models}
Token reduction has long been studied in vision transformers through importance-based pruning and similarity-based merging~\citep{liang2022evit,bolya2023tome}. Training-free VLM methods adapt these ideas to reduce visual sequences during inference. FastV prunes low-attention visual tokens in deeper LLM layers, whereas SparseVLM uses instruction-conditioned attention and adapts its layerwise sparsification ratio~\citep{chen2024fastv,zhang2024sparsevlm}. These attention-driven strategies prioritize evidence emphasized by the model or query, but can underrepresent broader image context under tight budgets. VisionZip retains dominant tokens and merges contextual tokens, while DivPrune selects a max-min diverse subset~\citep{yang2025visionzip,alvar2025divprune}. Their dominance- and diversity-based criteria preserve representative or nonredundant visual content, but do not explicitly prioritize evidence requested by the current query. MMTok brings relevance and visual coverage together through a submodular objective that covers both text and visual tokens with a fixed coefficient~\citep{dong2025mmtok}. This fixed scalarization applies the same relevance--coverage exchange rate at every selection state and retained-set size. StackTok instead keeps relevance as the objective and uses an entropy-adjusted, subset-size-indexed coverage reference to switch the active criterion within a single selection sequence.

\section{Method}

\subsection{Overview and Problem Setup}
StackTok selects a fixed-size visual-token subset before LLM inference. Figure~\ref{fig:framework} visualizes the method as a token-stacking trajectory. Post-projector query and visual embeddings yield query--vision affinities, while pre-projector vision features yield vision--vision affinities. A coverage-only greedy curve provides a size-indexed coverage reference, and query--vision affinity entropy calibrates the corresponding support target. The current subset's position relative to this target determines whether relevance or visual coverage extends the trajectory next. For high-resolution inputs, crop-local nominees feed one shared-budget trajectory, followed by optional support-preserving swap refinement. 

Let a VLM encode an image into $n\ge1$ aligned visual tokens. For a positive integer $N$, write $[N]=\{1,\ldots,N\}$, with the convention $[0]=\varnothing$. We denote the tokens' pre-projector vision features by $\mathbf{X}=[\mathbf{x}_1^\top;\ldots;\mathbf{x}_n^\top]\in\mathbb{R}^{n\times d_v}$ and the post-projector embeddings passed to the LLM by $\mathbf{Z}=[\mathbf{z}_1^\top;\ldots;\mathbf{z}_n^\top]\in\mathbb{R}^{n\times d}$. Here $d_v$ and $d$ are the vision-feature and LLM-embedding dimensions, respectively. After model-specific query preprocessing, the LLM input-embedding layer yields $m\ge0$ query-token embeddings $\mathbf{Q}=[\mathbf{q}_1^\top;\ldots;\mathbf{q}_m^\top]\in\mathbb{R}^{m\times d}$. The set $\mathcal{U}\subseteq[n]$ contains indices eligible for retention and excludes padding positions when present; all $n$ encoded positions remain affinity targets in Eq.~\eqref{eq:affinity}. Given a non-negative integer budget $K$, the goal is to retain exactly $k=\min\{K,|\mathcal{U}|\}$ tokens.

\subsection{Query Relevance and Visual Support}
We measure each signal in the representation suited to its role. Query relevance uses $\mathbf Q$ and $\mathbf Z$ in the shared LLM embedding space, whereas visual coverage uses $\mathbf X$, which preserves the vision encoder's geometry. Let $\bar{\mathbf{u}}=\mathbf{u}/\max(\|\mathbf{u}\|_2,\varepsilon_{\mathrm{norm}})$ denote $\ell_2$ normalization, where $\varepsilon_{\mathrm{norm}}>0$ prevents division by zero. Row-wise softmax then gives
\begin{equation}
\begin{aligned}
A^{qv}_{ij}
&=\frac{\exp(\bar{\mathbf q}_i^\top\bar{\mathbf z}_j/\tau_t)}
{\sum_{\ell=1}^{n}\exp(\bar{\mathbf q}_i^\top\bar{\mathbf z}_\ell/\tau_t)},\\
A^{vv}_{ij}
&=\frac{\exp(\bar{\mathbf x}_i^\top\bar{\mathbf x}_j/\tau_v)}
{\sum_{\ell=1}^{n}\exp(\bar{\mathbf x}_i^\top\bar{\mathbf x}_\ell/\tau_v)},
\end{aligned}
\label{eq:affinity}
\end{equation}
where the first line is indexed by $i\in[m]$ and $j\in[n]$, and the second by $i,j\in[n]$; $\tau_t,\tau_v>0$ control concentration. Here $\mathbf A^{qv}\in\mathbb R_+^{m\times n}$ maps query tokens to visual evidence, while $\mathbf A^{vv}\in\mathbb R_+^{n\times n}$ measures how well retained tokens represent the encoded image. Both are embedding affinities.

For any non-negative affinity matrix $\mathbf A{\in}\mathbb R_+^{p\times q}$ with $p,q\ge1$ and any $S\subseteq[q]$, define the facility-location coverage

\vspace{-1.2em}
\begin{equation}
c_i(S;\mathbf A)=\max_{j\in S}A_{ij},\qquad
F_{\mathbf A}(S)=\frac{1}{p}\sum_{i=1}^{p}c_i(S;\mathbf A),
\label{eq:coverage}
\end{equation}
with $c_i(\varnothing;\mathbf A)=0$. The maximum assigns each target its best representative in $S$, and the mean normalizes the number of target rows. We instantiate
\begin{equation}
R(S)=
\begin{cases}
F_{\mathbf A^{qv}}(S),&m{>}0,\\
0,&m{=}0,
\end{cases}
\qquad
C(S)=F_{\mathbf A^{vv}}(S),
\label{eq:relevance-support}
\end{equation}
as query relevance and visual coverage, respectively. For any
$S\subseteq[q]$ and $s\in[q]\setminus S$, the exact marginal gain of
$F_{\mathbf A}$ is
\begin{equation}
  \begin{aligned}
\Delta_{F_{\mathbf A}}(s\mid S)&=F_{\mathbf A}(S\cup\{s\})-F_{\mathbf A}(S)\\
&=\frac{1}{p}\sum_{i=1}^{p}[A_{is}-c_i(S;\mathbf A)]_+.
\end{aligned}
\label{eq:marginal}
\end{equation}
Here $[x]_+=\max\{x,0\}$. We denote the two instances by $\Delta_R$ and $\Delta_C$. Throughout, \emph{visual coverage} refers to the measurable quantity $C$, whereas \emph{visual support} refers to the requirement that this quantity fulfills in the selection policy.

\begin{stackpropbox}
\textbf{Proposition 1 (Coverage Structure).}
\textit{For every $\mathbf A\ge0$ with at least one row, $F_{\mathbf A}$ is normalized, monotone, and submodular, and its marginal gain is given by Eq.~\eqref{eq:marginal}.}
\end{stackpropbox}
Thus $R$ and $C$ have diminishing returns and admit exact incremental greedy updates~\citep{nemhauser1978,krause2014submodular}; when $m=0$, $R\equiv0$ has these properties trivially. The proof is in the supplementary appendix.

\subsection{Budget- and Input-Calibrated Visual Support}
A fixed objective $R(S)+\alpha C(S)$ with coefficient $\alpha\ge0$ uses the same relevance--coverage exchange rate across inputs, budgets, and selection states. StackTok instead derives a budget-specific coverage scale from a coverage-only greedy sequence initialized by $\mathcal G_0=\varnothing$. For $t=1,\ldots,k$,
\begin{equation}
\begin{aligned}
a_t&\in\argmax_{s\in\mathcal U\setminus\mathcal G_{t-1}}\Delta_C(s\mid\mathcal G_{t-1}),\\
\mathcal G_t&=\mathcal G_{t-1}\cup\{a_t\},\qquad C_t^{\mathrm{ref}}=C(\mathcal G_t),
\end{aligned}
\label{eq:reference}
\end{equation}
with $C_0^{\mathrm{ref}}=0$. StackTok discards the sets $\mathcal G_t$ and retains only the achieved coverage reference curve $\{C_t^{\mathrm{ref}}\}_{t=0}^{k}$.

\begin{stackpropbox}
\textbf{Theorem 1 (Anytime Coverage Reference).}
\textit{Let $\mathrm{OPT}_t=\max_{S\subseteq\mathcal U,\,|S|\le t}C(S)$. For every $1\le t\le k$,
\begin{equation}
C_t^{\mathrm{ref}}\ge\left[1{-}\left(1{-}\frac{1}{t}\right)^t\right]\mathrm{OPT}_t
\ge(1{-}e^{-1})\mathrm{OPT}_t.
\label{eq:reference-bound}
\end{equation}
Moreover, $C_t^{\mathrm{ref}}$ is nondecreasing and its increments $C_t^{\mathrm{ref}}-C_{t-1}^{\mathrm{ref}}$ are nonincreasing.}
\end{stackpropbox}
Thus $C_t^{\mathrm{ref}}$ is an attained, near-optimal same-budget scale with a diminishing-return shape, not an upper bound. For every $\beta\in[0,1]$, the same coverage-greedy chain meets the support target $\beta C_t^{\mathrm{ref}}$; this witnesses same-budget target feasibility, not feasibility of StackTok's interleaved prefixes. The proof is in the supplementary appendix.

The coverage reference sets the budget scale; query--vision affinity determines how closely to track it. For $m>0$ and $n>1$, StackTok computes each row entropy $H_i$ for $i\in[m]$, its normalized mean $\bar H$, and the support strictness $\beta$:
\begin{equation}
\begin{aligned}
H_i&=-\sum_{j=1}^{n}A^{qv}_{ij}\log A^{qv}_{ij},\\
\bar H&=\frac{1}{m\log n}\sum_{i=1}^{m}H_i,\\
\beta&=\operatorname{clip}(\bar H,\beta_{\min},\beta_{\max}),
\end{aligned}
\label{eq:beta}
\end{equation}
where the theoretical entropy uses the convention $0\log0=0$. In finite precision, $\delta>0$ is a numerical floor used only to evaluate the logarithm as $\log(\max\{A^{qv}_{ij},\delta\})$, preventing $\log0$ after underflow. Furthermore, $0\le\beta_{\min}\le\beta_{\max}\le1$, and $\operatorname{clip}$ truncates its argument to this interval. Low entropy permits greater emphasis on localized evidence, whereas high entropy requests broader visual support. Since diffusion may also arise from weak localization or alignment, $\beta$ calibrates the policy rather than classifying the query. When $m=0$, we bypass entropy calibration, set $\beta=\beta_{\max}$ for reference bookkeeping, and use visual coverage throughout. Otherwise, if $n=1$, we set $\beta=\beta_{\min}$; the single candidate makes selection unique. For $t\in\{0,\ldots,k\}$, the size-$t$ support target is
\begin{equation}
\theta_t=\beta C_t^{\mathrm{ref}}.
\label{eq:support-target}
\end{equation}

\subsection{Reference-Gated Interleaved Selection}
Initialize $S_0=\varnothing$. At each step $t\in\{0,\ldots,k-1\}$, let $S_t\subseteq\mathcal U$, with $|S_t|=t$, be the current retained set and let $\mathcal U_t=\mathcal U\setminus S_t$ be the remaining candidates. The support deficit is
\begin{equation}
d_t=[\theta_t-C(S_t)]_+.
\label{eq:deficit}
\end{equation}
For a nonempty query, zero deficit activates relevance and positive deficit activates visual coverage:
\begin{equation}
(g_t^{\mathrm{pri}},g_t^{\mathrm{sec}})=
\begin{cases}
(\Delta_R,\Delta_C),&d_t=0,\\
(\Delta_C,\Delta_R),&d_t>0.
\end{cases}
\label{eq:gate}
\end{equation}
StackTok uses the primary criterion while it has positive gain and otherwise falls back to the secondary one. Let $M_t^{\mathrm{pri}}=\max_{u\in\mathcal U_t}g_t^{\mathrm{pri}}(u\mid S_t)$. Then
\begin{equation}
\begin{aligned}
\psi_t(s)=
\begin{cases}
g_t^{\mathrm{pri}}(s\mid S_t),&M_t^{\mathrm{pri}}>0,\\
g_t^{\mathrm{sec}}(s\mid S_t),&\text{otherwise},
\end{cases}\\
s_{t+1}\in\argmax_{s\in\mathcal U_t}\psi_t(s).
\end{aligned}
\label{eq:selection}
\end{equation}
We then set $S_{t+1}=S_t\cup\{s_{t+1}\}$. When $m=0$, the coverage-oriented branch overrides Eq.~\eqref{eq:gate}. Deterministic tie-breaking fills the budget if both criteria saturate.

For any $s\in\mathcal U_t$ and fixed $\theta_t$, adding $s$ repairs the current deficit by exactly
\begin{equation}
d_t-[\theta_t-C(S_t\cup\{s\})]_+=\min\{d_t,\Delta_C(s\mid S_t)\},
\label{eq:deficit-repair}
\end{equation}
so the coverage-oriented branch maximizes one-step repair. Because the target advances to $\theta_{t+1}$ after selection, this local result does not certify the new prefix. Also, $\theta_0=C(\varnothing)=0$, so a nonempty query activates relevance as the primary criterion at the first step, subject to the zero-gain fallback in Eq.~\eqref{eq:selection}.

Algorithm~\ref{alg:single} instantiates the single-image trajectory in Figure~\ref{fig:framework}(a). Updating both coverages lets the gate switch repeatedly; sorting the final indices preserves the spatial order of the retained rows of $\mathbf Z$.

\begin{algorithm}[tb]
\caption{Reference-gated interleaved selection (single image)}
\label{alg:single}
\footnotesize
\textbf{Input}: affinities \algrel{$\mathbf A^{qv}$}, \algcov{$\mathbf A^{vv}$}; selectable set $\mathcal U$; budget $K$; strictness clipping range $[\beta_{\min},\beta_{\max}]$\\
\textbf{Output}: retained index set $S$
\begin{algorithmic}[1]
\STATE $k\leftarrow\min\{K,|\mathcal U|\}$; $S_0\leftarrow\varnothing$ \COMMENT{fixed-budget state}
\STATE compute \algcov{$\{C_t^{\mathrm{ref}}\}_{t=0}^{k}$} and $\beta$ \COMMENT{budget/input calibration}
\FOR{$t=0$ to $k-1$}
\STATE $d_t\leftarrow[\beta C_t^{\mathrm{ref}}-C(S_t)]_+$ \COMMENT{support feedback}
\STATE set $(g_t^{\mathrm{pri}},g_t^{\mathrm{sec}})$ by Eq.~\eqref{eq:gate}; use $(\Delta_C,\Delta_R)$ if $m=0$ \COMMENT{\algrel{relevance} / \algcov{coverage}}
\STATE $s_{t+1}\leftarrow\argmax_{s\in\mathcal U\setminus S_t}\psi_t(s)$ \COMMENT{fallback in $\psi_t$}
\STATE $S_{t+1}\leftarrow S_t\cup\{s_{t+1}\}$; update \algrel{$R$} and \algcov{$C$} \COMMENT{close feedback loop}
\ENDFOR
\STATE \textbf{return} $\operatorname{sort}(S_k)$ \COMMENT{restore spatial order}
\end{algorithmic}
\end{algorithm}

\subsection{Shared-Budget Allocation across Crops}
Figure~\ref{fig:framework}(b) visualizes the high-resolution extension. With $L\ge1$ crops indexed by $c\in[L]$, fixed per-crop quotas can waste tokens on uninformative regions. Crop $c$ contains $n_c\ge1$ encoded positions and a selectable set $\mathcal U_c\subseteq[n_c]$. The effective shared budget and crop-$c$ reference horizon are
$K_{\mathrm{tot}}=\min\{K,\sum_{c=1}^{L}|\mathcal U_c|\}$ and
$r_c^{\max}=\min\{K_{\mathrm{tot}},|\mathcal U_c|\}$, respectively.
Applying Eq.~\eqref{eq:affinity} within the crop gives $\mathbf A_c^{qv}\in\mathbb R_+^{m\times n_c}$ and $\mathbf A_c^{vv}\in\mathbb R_+^{n_c\times n_c}$. We define $C_c=F_{\mathbf A_c^{vv}}$ and, when $m>0$, $R_c=F_{\mathbf A_c^{qv}}$; when $m=0$, $R_c\equiv0$. Write $\Delta_{R_c}$ and $\Delta_{C_c}$ for their marginal gains. Each crop constructs $\{C_{c,t}^{\mathrm{ref}}\}_{t=0}^{r_c^{\max}}$ and $\beta_c$ from Eqs.~\eqref{eq:reference} and~\eqref{eq:beta}, replacing $(\mathbf A^{qv},\mathbf A^{vv},\mathcal U,n,k)$ by $(\mathbf A_c^{qv},\mathbf A_c^{vv},\mathcal U_c,n_c,r_c^{\max})$. The same edge-case conventions apply: $\beta_c=\beta_{\max}$ when $m=0$, while $\beta_c=\beta_{\min}$ when $m>0$ and $n_c=1$. For $t\in\{0,\ldots,r_c^{\max}\}$, define the local support target $\theta_{c,t}=\beta_c C_{c,t}^{\mathrm{ref}}$.

Initialize $S_c=\varnothing$ for every crop. At a shared-budget step, let $r_c=|S_c|$ and define the active-crop set $\mathcal I=\{c\in[L]:\mathcal U_c\setminus S_c\ne\varnothing\}$. Crop $c$ has support deficit $d_{c,r_c}=[\theta_{c,r_c}-C_c(S_c)]_+$. Applying the gate and zero-primary-gain fallback in Eqs.~\eqref{eq:gate}--\eqref{eq:selection} to $S_c$, $\mathcal U_c\setminus S_c$, $d_{c,r_c}$, $\Delta_{R_c}$, and $\Delta_{C_c}$ defines the crop-local score $\psi_{c,r_c}$. As in the single-image case, $m=0$ forces the coverage-oriented branch. Each $c\in\mathcal I$ then nominates
\begin{equation}
\begin{aligned}
s_c^\star&\in\argmax_{s\in\mathcal U_c\setminus S_c}\psi_{c,r_c}(s),\\
h_c&=\Delta_{R_c}(s_c^\star\mid S_c)+\Delta_{C_c}(s_c^\star\mid S_c).
\end{aligned}
\label{eq:crop-nominee}
\end{equation}
The allocator spends the $K_{\mathrm{tot}}$ slots by repeatedly choosing
\begin{equation}
\begin{aligned}
c^\star&\in\argmax_{c\in\mathcal I}h_c,\\
S_{c^\star}&\leftarrow S_{c^\star}\cup\{s_{c^\star}^\star\}.
\end{aligned}
\label{eq:global}
\end{equation}
The local gate determines each nominee, while $h_c$ compares nominees across crops. Only the winning crop changes, so other nominations are cached. Cross-crop allocation deliberately uses $\Delta_{R_c}+\Delta_{C_c}$; criterion separation applies to the local gate. Algorithm~\ref{alg:multi} makes these two levels explicit. All local and cross-crop maximizations use deterministic tie-breaking, and zero-gain nominees remain eligible so that the effective shared budget is filled.

\begin{algorithm}[tb]
\caption{Shared-budget allocation across crops}
\label{alg:multi}
\footnotesize
\textbf{Input}: per-crop affinities $\{\algrel{\mathbf A_c^{qv}},\algcov{\mathbf A_c^{vv}}\}_{c=1}^{L}$; selectable sets $\{\mathcal U_c\}_{c=1}^{L}$; shared budget $K$; strictness clipping range $[\beta_{\min},\beta_{\max}]$\\
\textbf{Output}: allocated index sets $\{S_c\}_{c=1}^{L}$
\begin{algorithmic}[1]
\STATE $K_{\mathrm{tot}}\leftarrow\min\{K,\sum_c|\mathcal U_c|\}$ \COMMENT{effective budget}
\STATE $S_c\leftarrow\varnothing$ for all $c$; $\mathcal I\leftarrow\{c:\mathcal U_c\ne\varnothing\}$ \COMMENT{active crops}
\FOR{$c\in\mathcal I$}
\STATE $r_c^{\max}\leftarrow\min\{K_{\mathrm{tot}},|\mathcal U_c|\}$ \COMMENT{local reference horizon}
\STATE compute \algcov{$\{C_{c,t}^{\mathrm{ref}}\}_{t=0}^{r_c^{\max}}$} and $\beta_c$ \COMMENT{local calibration}
\STATE cache $(s_c^\star,h_c)$ by Eq.~\eqref{eq:crop-nominee} \COMMENT{\algrel{relevance} / \algcov{coverage}}
\ENDFOR
\FOR{$b=1$ to $K_{\mathrm{tot}}$}
\STATE $c^\star\leftarrow\argmax_{c\in\mathcal I}h_c$ \COMMENT{compare only nominees}
\STATE $S_{c^\star}\leftarrow S_{c^\star}\cup\{s_{c^\star}^\star\}$; update \algrel{$R_{c^\star}$} and \algcov{$C_{c^\star}$} \COMMENT{update winner only}
\IF{$\mathcal U_{c^\star}\setminus S_{c^\star}=\varnothing$}
\STATE $\mathcal I\leftarrow\mathcal I\setminus\{c^\star\}$ \COMMENT{crop exhausted}
\ELSE
\STATE refresh $(s_{c^\star}^\star,h_{c^\star})$ by Eq.~\eqref{eq:crop-nominee} \COMMENT{cache other nominees}
\ENDIF
\ENDFOR
\STATE \textbf{return} $\{\operatorname{sort}(S_c)\}_{c=1}^{L}$ \COMMENT{restore within-crop order}
\end{algorithmic}
\end{algorithm}

\subsection{Support-Preserving Swap Refinement}
Greedy additions cannot revise early choices. Let $\mathcal U_{\mathrm{loc}}$, $R_{\mathrm{loc}}$, $C_{\mathrm{loc}}$, $\beta_{\mathrm{loc}}$, and $k_{\mathrm{loc}}$ denote the quantities of the unit being refined. For a single image,
$(\mathcal U_{\mathrm{loc}},R_{\mathrm{loc}},C_{\mathrm{loc}},\beta_{\mathrm{loc}},k_{\mathrm{loc}})
=(\mathcal U,R,C,\beta,k)$ and
$C_{\mathrm{loc},t}^{\mathrm{ref}}=C_t^{\mathrm{ref}}$ for $0\le t\le k$.
For crop $c$, they are
$(\mathcal U_c,R_c,C_c,\beta_c,r_c^{\max})$ and
$C_{\mathrm{loc},t}^{\mathrm{ref}}=C_{c,t}^{\mathrm{ref}}$ for
$0\le t\le r_c^{\max}$. Given a retained set
$S\subset\mathcal U_{\mathrm{loc}}$ with
$0<r=|S|\le k_{\mathrm{loc}}$ and
$r<|\mathcal U_{\mathrm{loc}}|$, StackTok optionally performs cardinality-preserving one-exchange refinement. It evaluates replacements $S'=S\setminus\{u\}\cup\{v\}$ with $u\in S$ and $v\in\mathcal U_{\mathrm{loc}}\setminus S$. Define $J_{\mathrm{loc}}(S)=R_{\mathrm{loc}}(S)+C_{\mathrm{loc}}(S)$. A replacement is accepted only if
\begin{equation}
\begin{aligned}
J_{\mathrm{loc}}(S')&>(1+\epsilon_{\mathrm{sw}})J_{\mathrm{loc}}(S),\\
C_{\mathrm{loc}}(S')&\ge\ell_r(S):=
\min\{\beta_{\mathrm{loc}}C_{\mathrm{loc},r}^{\mathrm{ref}},
C_{\mathrm{loc}}(S)\}.
\end{aligned}
\label{eq:swap}
\end{equation}
Here $\epsilon_{\mathrm{sw}}\ge0$ is the minimum relative improvement. Below target, the floor prevents support loss; above target, it permits a decrease only down to the target. Each accepted replacement preserves $r$ and strictly increases $J_{\mathrm{loc}}$. Because the family of size-$r$ subsets is finite, repeated accepted replacements must terminate. Per-row top-two affinities make deletion scores exact without recomputing coverage; in practice, refinement stops after a fixed number of passes or a full pass without an accepted swap.

\begin{table*}[t]
\centering
\footnotesize
\stacktablestyle
\setlength{\tabcolsep}{3.5pt}
\renewcommand{\arraystretch}{0.95}
\begin{tabular}{lcccccccccc}
\toprule[0.9pt]
\stacktableheadrow
\textbf{Method} & \textbf{GQA}$\uparrow$ & \textbf{MMB}$\uparrow$ & \textbf{MME}$\uparrow$ & \textbf{POPE}$\uparrow$ & \textbf{SQA}$\uparrow$ & \textbf{\vqavtwo}$\uparrow$ & \textbf{\vqatext}$\uparrow$ & \textbf{MMMU}$\uparrow$ & \textbf{SEED}$\uparrow$ & \textbf{Avg.}$\uparrow$ \\
\midrule[0.45pt]
\stacktablebaserow
Vanilla (576) & 61.95 & 64.18 & 1861.48 & 85.87 & 69.51 & 77.71 & 58.15 & 36.22 & 58.55 & 100.00\% \\
\midrule[0.45pt]
\stacktablegrouprow
\multicolumn{11}{c}{\textit{\textbf{Retain 192}}\quad\stacktableaccent{\textit{\textbf{($\downarrow 67\%$)}}}}\\
FastV & 52.74 & 60.71 & 1611.55 & 64.78 & 67.31 & 66.42 & 52.45 & 34.22 & 57.05 & 89.57\% \\
SparseVLM & 57.65 & 62.00 & 1720.52 & 83.57 & 69.11 & 74.84 & 56.05 & 33.73 & 55.75 & 95.54\% \\
VisionZip & 59.35 & 62.49 & 1782.10 & 85.27 & 68.91 & 76.03 & 57.25 & 36.52 & 56.35 & 97.86\% \\
DivPrune & 60.02 & 62.04 & 1761.74 & 86.97 & 68.67 & 76.10 & 56.92 & 35.36 & 58.66 & 97.99\% \\
VisionZip$^\ddagger$ & 60.15 & 62.89 & 1833.49 & 84.87 & 68.21 & 76.62 & 57.75 & 36.12 & 57.05 & 98.40\% \\
                   MMTok & 60.12 & 62.89 & 1773.36 & 86.39 & 68.77 & 76.33 & 57.63 & 36.25 & 59.16 & 98.70\% \\
\stacktableoursrow
\textbf{StackTok (ours)} & 60.32 & 63.77 & 1768.84 & 86.70 & 69.37 & 76.54 & 57.49 & 36.22 & 59.16 & \textbf{98.99\%} \\
\midrule
\addlinespace[1pt]
\stacktablegrouprow
\multicolumn{11}{c}{\textit{\textbf{Retain 128}}\quad\stacktableaccent{\textit{\textbf{($\downarrow 78\%$)}}}}\\
FastV & 49.64 & 55.65 & 1489.58 & 59.58 & 60.21 & 61.18 & 50.56 & 34.82 & 55.85 & 84.45\% \\
SparseVLM & 56.05 & 59.52 & 1695.53 & 80.47 & 67.11 & 73.06 & 54.85 & 33.73 & 53.35 & 93.02\% \\
VisionZip & 57.65 & 61.50 & 1761.21 & 83.17 & 68.91 & 74.84 & 56.75 & 37.82 & 54.85 & 96.83\% \\
DivPrune & 59.30 & 61.53 & 1717.74 & 86.69 & 68.67 & 75.20 & 56.01 & 35.48 & 56.93 & 96.88\% \\
VisionZip$^\ddagger$ & 58.95 & 62.10 & 1822.49 & 83.67 & 68.31 & 75.83 & 56.95 & 37.22 & 55.75 & 97.67\% \\
                   MMTok & 59.34 & 61.79 & 1778.64 & 86.22 & 68.83 & 75.58 & 56.98 & 35.59 & 58.54 & 97.84\% \\
\stacktableoursrow
\textbf{StackTok (ours)} & 59.52 & 62.80 & 1761.15 & 86.47 & 69.31 & 75.59 & 56.84 & 35.56 & 58.60 & \textbf{98.03\%} \\
\midrule
\addlinespace[1pt]
\stacktablegrouprow
\multicolumn{11}{c}{\textit{\textbf{Retain 64}}\quad\stacktableaccent{\textit{\textbf{($\downarrow 89\%$)}}}}\\
FastV & 46.14 & 47.61 & 1255.65 & 47.98 & 51.11 & 54.45 & 47.76 & 33.93 & 51.86 & 75.55\% \\
SparseVLM & 52.74 & 55.75 & 1504.58 & 75.07 & 62.21 & 67.51 & 51.76 & 32.63 & 51.06 & 86.99\% \\
VisionZip & 55.14 & 59.62 & 1689.53 & 76.97 & 69.01 & 71.67 & 55.45 & 36.12 & 52.16 & 93.11\% \\
DivPrune & 57.83 & 58.80 & 1673.93 & 85.53 & 68.08 & 73.36 & 54.64 & 35.48 & 55.08 & 94.76\% \\
VisionZip$^\ddagger$ & 57.05 & 61.01 & 1755.51 & 80.87 & 68.81 & 73.45 & 55.95 & 35.52 & 53.35 & 94.95\% \\
                   MMTok & 58.34 & 60.68 & 1714.85 & 85.74 & 69.17 & 74.44 & 55.96 & 36.03 & 57.10 & 96.58\% \\
\stacktableoursrow
\textbf{StackTok (ours)} & 58.42 & 61.25 & 1702.47 & 85.89 & 69.65 & 74.44 & 55.85 & 36.00 & 56.99 & \textbf{96.66\%} \\
\bottomrule[0.9pt]
\end{tabular}
\vspace{-0.3em}
\caption{Performance on LLaVA-1.5-7B over nine benchmarks; Avg. is the mean relative retention with respect to the 576-token Vanilla row and is recomputed from the displayed values after rounding. $\ddagger$ denotes the fine-tuned variant.
}
\vspace{-1em}
\label{tab:main1}
\end{table*}

\begin{table}[t]
\centering
\footnotesize
\stacktablestyle
\setlength{\tabcolsep}{4pt}
\begin{tabular}{lccc}
\toprule[0.9pt]
\stacktableheadrow
\textbf{Variant} & \textbf{192} & \textbf{128} & \textbf{64} \\
\midrule[0.45pt]
(a) Final-budget target ($C_t^{\mathrm{ref}}\!\to C_k^{\mathrm{ref}}$) & 98.70 & 97.58 & 96.05 \\
(b) Fixed-strictness control ($\beta=0.5$) & 98.88 & 97.92 & 96.54 \\
(c) Relevance only ($\psi_t=\Delta_R$) & 98.10 & 96.75 & 94.20 \\
(d) Coverage only ($\psi_t=\Delta_C$) & 98.15 & 96.95 & 94.55 \\
\stacktableoursrow
\textbf{StackTok (full)} & 98.99 & 98.03 & 96.66 \\
\bottomrule[0.9pt]
\end{tabular}
\vspace{-0.3em}
\caption{Ablations of the size-indexed coverage reference, entropy calibration, and active criterion on LLaVA-1.5-7B (average retained performance, \%).}
\vspace{-1.5em}
\label{tab:ablate-component}
\end{table}

\subsection{Complexity and Practical Details}
Affinity construction costs $O(n^2d_v+mnd)$ time and $O(n^2+mn)$ memory. For $u=|\mathcal U|$, incremental row maxima make the reference and interleaved passes cost $O(ku(n+m))$, or $O(kn^2)$ when $u,m\le n$. For refinement, let $u_{\mathrm{loc}}=|\mathcal U_{\mathrm{loc}}|$ and let $n_{\mathrm{loc}}=n$ for a single image or $n_{\mathrm{loc}}=n_c$ for crop $c$; one complete swap pass then costs $O(r(u_{\mathrm{loc}}-r)(m+n_{\mathrm{loc}}))$. Multi-crop costs sum over crops, with cached unchanged nominations. Without query embeddings, StackTok uses visual coverage throughout. It excludes padding from selection, does not stop early, and returns the effective fixed budget.
\section{Experiments}

\subsection{Experimental Setup}
\noindent\textbf{Datasets, models, and metrics.}
The evaluation covers ten distinct image-understanding benchmarks. The LLaVA suite contains GQA~\citep{gqa}, MMBench (MMB)~\citep{mmbench}, MME~\citep{mme}, POPE~\citep{pope}, ScienceQA-IMG (SQA)~\citep{sqa}, \vqavtwo~\citep{vqav2}, \vqatext~\citep{textvqa}, MMMU~\citep{mmmu}, and SEEDBench (SEED)~\citep{seed}. We evaluate LLaVA-1.5-7B/13B and LLaVA-NeXT-7B/13B~\citep{liu2023llava,liu2024llavanext} using the lmms-eval framework~\citep{lmmseval}. Tables~\ref{tab:main1} and~\ref{tab:crossmodel} report the average relative retention across the applicable datasets, computed as each pruned score divided by its corresponding full-token score. We additionally evaluate Qwen2.5-VL-7B~\citep{bai2025qwenvl} on GQA, MMB, MME, POPE, SQA, \vqatext, and OCRBench~\citep{liu2024ocrbench}. Due to space constraints, the Qwen2.5-VL-7B experiments are presented in the supplementary appendix. For these experiments, Avg.$^\dagger$ averages relative retention over GQA, MMB, MME, POPE, and \vqatext, while SQA and OCRBench are excluded from the aggregate and reported separately. Extended ablations, implementation details, and limitations are also provided in the appendix.

\noindent\textbf{Baselines and budgets.}
We compare with the training-free token-reduction methods FastV~\citep{chen2024fastv}, SparseVLM~\citep{zhang2024sparsevlm}, VisionZip~\citep{yang2025visionzip}, DivPrune~\citep{alvar2025divprune}, and MMTok~\citep{dong2025mmtok}. VisionZip$^\ddagger$ denotes its fine-tuned variant and is included only as an additional reference; it is excluded when identifying the strongest training-free baseline. For LLaVA-1.5, we retain 192/128/64 of the original 576 visual tokens. For LLaVA-NeXT, we apply a shared budget of 640/320/160 visual tokens across all crops of an input. For the Qwen2.5-VL-7B experiments in the appendix, we retain 20\%/10\%/5\% of the visual tokens produced by dynamic-resolution encoder.

\begingroup
\let\stackincludegraphics\includegraphics
\renewcommand{\includegraphics}[2][]{\stackincludegraphics[clip,#1]{#2}}

\begin{figure*}[t]
\centering
\includegraphics[width=\textwidth]{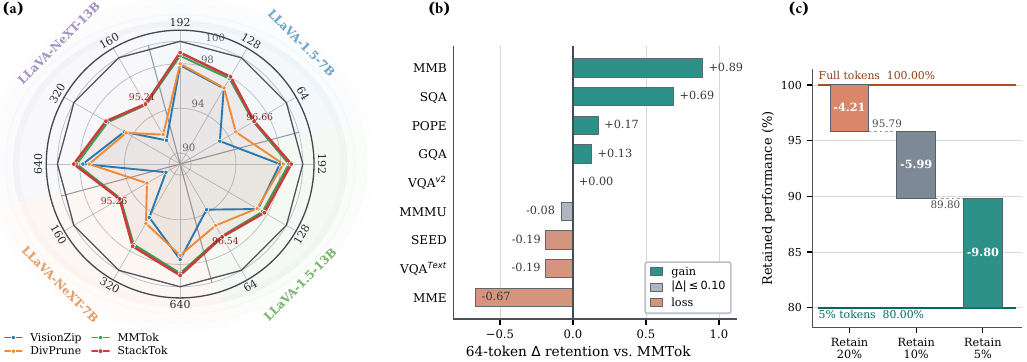}
\vspace{-1.5em}
\caption{Complementary views of the main results. (a) Dataset-averaged retained performance for four training-free methods, reported separately for each LLaVA model and tested budget in Table~\ref{tab:crossmodel}; the radial axis is truncated to 89--101\%, and the four annotations mark StackTok at each model's tightest budget. (b) Per-benchmark StackTok-minus-MMTok difference on LLaVA-1.5-7B at the tightest 64-token budget (11.1\% of 576), after normalizing each displayed score by its Vanilla score in Table~\ref{tab:main1}. The near-zero category, defined by $\lvert\Delta\rvert\leq0.10$ percentage points, is a visualization rule rather than a statistical-significance threshold. (c) Successive changes in StackTok Avg.$^\dagger$ as the Qwen2.5-VL-7B retention ratio tightens in Table~\ref{tab:qwen}. Avg.$^\dagger$ covers GQA, MMB, MME, POPE, and \vqatext; the floating bars describe adjacent-budget changes rather than causal contributions.}
\vspace{-0.3em}
\label{fig:main-visuals}
\end{figure*}

\endgroup
\begin{table*}[t]
\centering
\footnotesize
\stacktablestyle
\setlength{\tabcolsep}{5.5pt}
\renewcommand{\arraystretch}{0.95}
\begin{tabular}{lcccc} 
\toprule[0.9pt]
\stacktableheadrow
\textbf{Method} & \textbf{LLaVA-1.5-7B} & \textbf{LLaVA-1.5-13B} & \textbf{LLaVA-NeXT-7B} & \textbf{LLaVA-NeXT-13B} \\
\stacktableheadrow
 & \stacktableaccent{\textbf{192/128/64}} & \stacktableaccent{\textbf{192/128/64}} & \stacktableaccent{\textbf{640/320/160}} & \stacktableaccent{\textbf{640/320/160}} \\
\midrule[0.45pt]
VisionZip & 97.86 / 96.83 / 93.11 & 97.94 / 97.06 / 93.72 & 97.56 / 94.53 / 90.47 & 97.72 / 94.78 / 91.45 \\
DivPrune & 97.99 / 96.88 / 94.76 & 98.24 / 96.97 / 95.36 & 97.18 / 95.14 / 92.43 & 97.16 / 94.59 / 92.05 \\
VisionZip$^\ddagger$ & 98.40 / 97.67 / 94.95 & 98.76 / 97.48 / 94.83 & 98.95 / 97.64 / 95.07 & 98.82 / 97.86 / 94.69 \\
MMTok & 98.70 / 97.84 / 96.58 & 98.75 / 97.53 / 96.46 & 98.78 / 97.32 / 95.18 & 98.27 / 96.49 / 95.13 \\
\stacktableoursrow
\textbf{StackTok (ours)} & \textbf{98.99 / 98.03 / 96.66} & \textbf{98.96 / 97.72 / 96.54} & \textbf{98.97 / 97.51 / 95.26} & \textbf{98.53 / 96.68 / 95.21} \\
\bottomrule[0.9pt]
\end{tabular}
\vspace{-0.5em}
\caption{Dataset-averaged retained performance (\%) for four LLaVA models, reported separately at each of three budgets. The LLaVA-1.5-7B column is aligned with Table~\ref{tab:main1}.}
\vspace{-1.7em}
\label{tab:crossmodel}
\end{table*}

\subsection{Main Results}
\noindent\textbf{Cross-model comparison.}
Table~\ref{tab:main1} gives the full LLaVA-1.5-7B comparison, where StackTok exceeds the strongest training-free baseline by 0.29, 0.19, and 0.08 points at 192, 128, and 64 tokens. Table~\ref{tab:crossmodel} extends this comparison to four LLaVA models: StackTok ranks first in all 12 model--budget settings, with gains of 0.08--0.29 points. The consistent ordering across 7B/13B scales and fixed-resolution/multi-crop inputs suggests that the gain is not tied to one model size or image encoding regime; Figure~\ref{fig:main-visuals}(a) visualizes this pattern. At the tightest 64-token budget, the largest gains over MMTok occur on MMB and SQA. This pattern may arise because these question-driven benchmarks reward concentrating the tight budget on query-aligned evidence, while the coverage gate preserves complementary visual support.

\noindent\textbf{High-resolution performance.}
The high-resolution LLaVA-NeXT-7B setting retains at most 2{,}880 visual tokens before selection. StackTok preserves 98.97\%, 97.51\%, and 95.26\% of full-token performance with shared budgets of 640, 320, and 160 tokens, corresponding to only 22.2\%, 11.1\%, and 5.6\% of the original sequence. The modest degradation after removing 94.4\% of the tokens suggests that crop-local nominations preserve support within each crop while the allocator redirects slots away from low-gain crops. LLaVA-NeXT-13B follows the same ranking, reaching 98.53\%, 96.68\%, and 95.21\%, which indicates that the behavior persists across model scales.

\noindent\textbf{Dynamic-resolution transfer.}
Figure~\ref{fig:main-visuals}(c) provides a complementary view on Qwen2.5-VL-7B: Avg.$^\dagger$ decreases from 95.79\% to 89.80\% and 80.00\% as retention halves from 20\% to 10\% and 5\%. The larger 9.80-point second drop suggests that visual redundancy becomes substantially scarcer below 10\% retention; detailed results are provided in the appendix.

\noindent\textbf{Task-level behavior.}
Figure~\ref{fig:main-visuals}(b) also exposes the boundary of the aggregate gain: at 64 tokens, StackTok trails MMTok on MME, \vqatext, and SEED. These tasks can depend on dispersed scene cues or fine-grained text; under extreme compression, such evidence may receive weak individual relevance before its joint value becomes apparent. A better average at a fixed budget therefore does not imply uniform per-benchmark dominance.

\vspace{-0.8em}
\subsection{Component ablation}
\vspace{-0.2em}
Table~\ref{tab:ablate-component} isolates the three decisions central to StackTok. The deficit from replacing the size-indexed reference with the final-budget target grows from 0.29 to 0.61 points as the budget tightens. This widening gap indicates that an endpoint-only target over-constrains early prefixes, forcing scarce slots toward coverage before enough query-relevant evidence has been collected. Fixed strictness yields a smaller but consistent 0.11--0.12-point loss because one global \(\beta\) cannot adapt the support requirement to input ambiguity. The one-signal variants expose the complementary failure modes more directly: relevance alone can select redundant query-aligned evidence, whereas coverage alone can preserve diverse yet task-irrelevant regions. Their larger losses at 64 tokens therefore support using the support-deficit gate to switch between the two criteria.

\vspace{-1.5em}
\section{Conclusion}
\vspace{-0.5em}
Visual-token selection under tight budgets is not simply a matter of maximizing a fixed combination of importance signals: query relevance and visual coverage play asymmetric, state-dependent roles. We introduced StackTok, which pursues query-relevant evidence while treating visual coverage as budget-calibrated support. A size-indexed coverage reference and input-adaptive strictness determine which criterion guides each selection step, while crop-local nominations extend the same principle to shared-budget high-resolution inputs. StackTok provides a training-free framework in which selection priorities adapt to the input and the evolving retained set rather than remaining fixed throughout pruning.

\bibliography{stacktok_arxiv}

\newpage
\appendix
\twocolumn[
\begin{center}
{\LARGE\bfseries Supplementary Material for\par}
\vspace{0.3em}
{\Large\bfseries StackTok: Accelerating VLMs Inference with Budget-Adaptive Visual Token Selection\par}
\end{center}
\vspace{0.5em}
]

\section*{Appendix Overview}
This appendix complements the main paper with complete proofs and additional experimental evidence. It first proves Proposition~1 and Theorem~1, then provides implementation details, dynamic-resolution results on Qwen2.5-VL-7B, and extended ablations of fixed scalarization, shared-budget crop allocation, support-preserving swap refinement, and strictness sensitivity.

\setcounter{table}{0}
\renewcommand{\thetable}{A\arabic{table}}

\section{Proofs}

\subsection{Proof of Proposition 1 (Coverage Structure)}
For completeness, recall that for a non-negative affinity matrix
$\mathbf A\in\mathbb R_+^{p\times q}$ with $p,q\ge1$,
\begin{equation*}
\begin{aligned}
c_i(S;\mathbf A)&=\max_{j\in S}A_{ij},\qquad
c_i(\varnothing;\mathbf A)=0,\\
F_{\mathbf A}(S)&=\frac{1}{p}\sum_{i=1}^{p}c_i(S;\mathbf A).
\end{aligned}
\end{equation*}

\textcolor{stackrelevance}{\textbf{Proposition 1 (Coverage Structure, Restated).}}\enspace
\textit{$F_{\mathbf A}$ is normalized, monotone, and submodular. Moreover, for every $S\subseteq[q]$ and $s\in[q]\setminus S$,
\begin{equation*}
\Delta_{F_{\mathbf A}}(s\mid S)
=\frac{1}{p}\sum_{i=1}^{p}
[A_{is}-c_i(S;\mathbf A)]_+.
\end{equation*}}

\noindent\textbf{Proof.}
We establish the four claims separately. First, the empty-set convention above gives
$F_{\mathbf A}(\varnothing)=0$, so the function is normalized. Next, if
$S\subseteq T$, then every index available in the maximum defining
$c_i(S;\mathbf A)$ is also available for $c_i(T;\mathbf A)$. Hence
\begin{equation*}
c_i(S;\mathbf A)\le c_i(T;\mathbf A)
\quad\text{for every }i,
\end{equation*}
including $S=\varnothing$ because $\mathbf A$ is non-negative. Averaging over
rows yields $F_{\mathbf A}(S)\le F_{\mathbf A}(T)$ and proves
monotonicity.

For any candidate $s\notin S$, adding $s$ changes the coverage of row $i$ by
\begin{equation*}
\begin{aligned}
&c_i(S\cup\{s\};\mathbf A)-c_i(S;\mathbf A)\\
&\quad=\max\{c_i(S;\mathbf A),A_{is}\}
       -c_i(S;\mathbf A)\\
&\quad=[A_{is}-c_i(S;\mathbf A)]_+.
\end{aligned}
\end{equation*}
Averaging this row-wise identity gives the stated marginal-gain formula.

Finally, let $S\subseteq T$ and $s\notin T$. The monotonicity just proved gives
$c_i(S;\mathbf A)\le c_i(T;\mathbf A)$, while
$x\mapsto[A_{is}-x]_+$ is nonincreasing. Therefore
\begin{equation*}
[A_{is}-c_i(S;\mathbf A)]_+
\ge [A_{is}-c_i(T;\mathbf A)]_+.
\end{equation*}
Summing this inequality over rows and invoking the stated marginal-gain formula yields
$\Delta_{F_{\mathbf A}}(s\mid S)\ge
\Delta_{F_{\mathbf A}}(s\mid T)$. This is the diminishing-returns
characterization of submodularity on the finite ground set.

\subsection{Proof of Theorem 1 (Anytime Coverage Reference)}
Let $C=F_{\mathbf A^{vv}}$ and consider the coverage-greedy chain
$\{\mathcal G_i\}_{i=0}^{k}$ in Eq.~\eqref{eq:reference}. Restricting
$C$ to $2^{\mathcal U}$ preserves normalization, monotonicity, and
submodularity. Moreover, because $k\le|\mathcal U|$, the candidate set is
nonempty at every greedy step $0\le i<k$.

{\textbf{Theorem 1 (Anytime Coverage Reference, Restated).}}\enspace
\textit{Let $\mathrm{OPT}_t=\max_{S\subseteq\mathcal U,\,|S|\le t}C(S)$. For every $1\le t\le k$,
\begin{equation*}
C_t^{\mathrm{ref}}\ge
\left[1-\left(1-\frac{1}{t}\right)^t\right]\mathrm{OPT}_t
\ge(1-e^{-1})\mathrm{OPT}_t.
\end{equation*}
Furthermore, $C_t^{\mathrm{ref}}$ is nondecreasing and its increments are nonincreasing.}

\begin{table*}[ht]
\centering
\footnotesize
\stacktablestyle
\setlength{\tabcolsep}{5.5pt}
\renewcommand{\arraystretch}{0.95}
\begin{tabular}{lcccccccc}
\toprule[0.9pt]
\stacktableheadrow
\textbf{Method} & \textbf{GQA}$\uparrow$ & \textbf{MMB}$\uparrow$ & \textbf{MME}$\uparrow$ & \textbf{POPE}$\uparrow$ & \textbf{\vqatext}$\uparrow$ & \textbf{SQA}$\uparrow$ & \textbf{OCRBench}$\uparrow$ & \textbf{Avg.$^\dagger$}$\uparrow$ \\
\midrule[0.45pt]
\stacktablebaserow
Vanilla & 60.48 & 83.25 & 2327 & 86.16 & 77.72 & 87.46 & 83.80 & 100.00\% \\
\midrule[0.45pt]
\stacktablegrouprow
\multicolumn{9}{c}{\textit{\textbf{Retain 20\%}}}\\
VisionZip & 56.80 & 80.33 & 2174 & 83.38 & 70.43 & 84.23 & 59.50 & 94.25\% \\
DivPrune & 56.70 & 76.98 & 2163 & 80.59 & 65.86 & 80.91 & 48.10 & 91.49\% \\
MMTok & 58.09 & 79.30 & 2217 & 82.38 & 70.49 & 81.61 & 59.60 & \textbf{94.58\%} \\
\stacktableoursrow
\textbf{StackTok (ours)} & 58.84 & 81.19 & 2232 & 83.48 & 71.00 & 82.32 & 59.78 & \textbf{95.79\%} \\
\midrule
\addlinespace[1pt]
\stacktablegrouprow
\multicolumn{9}{c}{\textit{\textbf{Retain 10\%}}}\\
VisionZip & 52.47 & 75.60 & 2003 & 78.90 & 63.78 & 82.30 & 36.90 & 87.46\% \\
DivPrune & 53.43 & 72.85 & 1957 & 74.99 & 59.59 & 79.57 & 37.30 & 84.73\% \\
MMTok & 55.09 & 74.74 & 2051 & 78.75 & 63.90 & 80.47 & 43.60 & \textbf{88.52\%} \\
\stacktableoursrow
\textbf{StackTok (ours)} & 55.94 & 76.89 & 2056 & 79.95 & 64.52 & 81.03 & 43.68 & \textbf{89.80\%} \\
\midrule
\addlinespace[1pt]
\stacktablegrouprow
\multicolumn{9}{c}{\textit{\textbf{Retain 5\%}}}\\
VisionZip & 46.28 & 67.53 & 1677 & 66.38 & 54.49 & 79.57 & 19.70 & 75.37\% \\
DivPrune & 49.01 & 65.89 & 1739 & 68.45 & 52.02 & 77.05 & 24.90 & 76.26\% \\
MMTok & 50.66 & 65.89 & 1796 & 71.35 & 55.95 & 77.19 & 30.70 & \textbf{78.98\%} \\
\stacktableoursrow
\textbf{StackTok (ours)} & 51.35 & 67.32 & 1805 & 72.34 & 56.52 & 77.73 & 30.73 & \textbf{80.00\%} \\
\bottomrule[0.9pt]
\end{tabular}
\caption{Performance on Qwen2.5-VL-7B under dynamic resolution. Avg.$^\dagger$ is the mean relative retention over GQA, MMB, MME, POPE, and \vqatext; SQA and OCRBench are excluded.}
\label{tab:qwen}
\end{table*}

\noindent\textbf{Proof.}
For every $0\le i<k$, define the gain of the next greedy step by
\begin{equation*}
\begin{aligned}
\eta_{i+1}
&=C(\mathcal G_{i+1})-C(\mathcal G_i)\\
&=\max_{s\in\mathcal U\setminus \mathcal G_i}\Delta_C(s\mid \mathcal G_i).
\end{aligned}
\end{equation*}
Fix any $t\in\{1,\ldots,k\}$, and let $S_t^\star$ attain
$\mathrm{OPT}_t$; such a set exists because $\mathcal U$ is finite. For
each $0\le i<t$, monotonicity first allows us to adjoin $\mathcal G_i$ to an optimal set.
Submodularity then bounds the joint gain by the sum of singleton marginal
gains evaluated at $\mathcal G_i$:
\begin{equation*}
\begin{aligned}
\mathrm{OPT}_t-C(\mathcal G_i)
&=C(S_t^\star)-C(\mathcal G_i)\\
&\le C(\mathcal G_i\cup S_t^\star)-C(\mathcal G_i)\\
&\le\sum_{s\in S_t^\star\setminus \mathcal G_i}
\Delta_C(s\mid \mathcal G_i)\\
&\le |S_t^\star\setminus \mathcal G_i|\,\eta_{i+1}
\le t\eta_{i+1}.
\end{aligned}
\end{equation*}
The second inequality can be seen by adding the elements of
$S_t^\star\setminus \mathcal G_i$ one at a time: diminishing returns makes the
actual marginal at each enlarged intermediate set no greater than its
marginal at $\mathcal G_i$. The next inequality follows from the definition of \(\eta_{i+1}\) above.

Define the same-budget optimality gap
$\Gamma_i^{(t)}=\mathrm{OPT}_t-C(\mathcal G_i)$. Since $|\mathcal G_i|=i\le t$, every $\mathcal G_i$ considered
above is feasible for $\mathrm{OPT}_t$, so $\Gamma_i^{(t)}\ge0$. Combining the preceding bound with
$\Gamma_{i+1}^{(t)}=\Gamma_i^{(t)}-\eta_{i+1}$ gives the contraction
\begin{equation*}
\Gamma_{i+1}^{(t)}
\le\left(1-\frac{1}{t}\right)\Gamma_i^{(t)}.
\end{equation*}
Because $C(\mathcal G_0)=0$, we have
$\Gamma_0^{(t)}=\mathrm{OPT}_t$. Iterating this contraction for $i=0,\ldots,t-1$ yields
\begin{equation*}
\mathrm{OPT}_t-C(\mathcal G_t)
\le\left(1-\frac{1}{t}\right)^t\mathrm{OPT}_t,
\end{equation*}
which gives the first claimed inequality because
$C(\mathcal G_t)=C_t^{\mathrm{ref}}$. For $t>1$, the elementary bound
$\log(1-x)\le-x$ at $x=1/t$ gives
$(1-1/t)^t\le e^{-1}$; the case $t=1$ follows directly. This proves the
approximation guarantee for every recorded prefix of the same greedy chain.

It remains to establish the shape of the reference curve. By monotonicity,
$\eta_i\ge0$, so $C_t^{\mathrm{ref}}$ is nondecreasing. For every
$1\le i<k$ and $s\in\mathcal U\setminus \mathcal G_i$, diminishing returns and
$\mathcal G_{i-1}\subseteq \mathcal G_i$ imply
\begin{equation*}
\begin{aligned}
\Delta_C(s\mid \mathcal G_i)
&\le\Delta_C(s\mid \mathcal G_{i-1})\\
&\le\max_{u\in\mathcal U\setminus \mathcal G_{i-1}}
\Delta_C(u\mid \mathcal G_{i-1})\\
&=\eta_i.
\end{aligned}
\end{equation*}
Taking the maximum over the smaller candidate set
$\mathcal U\setminus \mathcal G_i$ shows that $\eta_{i+1}\le\eta_i$; hence the
reference increments are nonincreasing.

\begin{table}[t]
\centering
\footnotesize
\stacktablestyle
\setlength{\tabcolsep}{4.5pt}
\begin{tabular}{lccc}
\toprule[0.9pt]
\stacktableheadrow
\textbf{Variant} & \textbf{192} & \textbf{128} & \textbf{64} \\
\midrule[0.45pt]
Fixed scalarization, best $\alpha$ & 98.85 & 97.93 & 96.60 \\
\stacktableoursrow
\textbf{Reference-gated policy (ours)} & 98.99 & 98.03 & 96.66 \\
\bottomrule[0.9pt]
\end{tabular}
\caption{Reference-gated policy versus the best per-budget fixed scalarization on LLaVA-1.5-7B (average retained performance, \%).}
\label{tab:ablate-constraint}
\end{table}

\begin{table}[t]
\centering
\footnotesize
\stacktablestyle
\setlength{\tabcolsep}{5pt}
\begin{tabular}{lccc}
\toprule[0.9pt]
\stacktableheadrow
\textbf{Allocation} & \textbf{640} & \textbf{320} & \textbf{160} \\
\midrule[0.45pt]
Uniform fixed quotas & 98.55 & 96.95 & 94.40 \\
Initial-$h_c$-proportional quotas & 98.72 & 97.20 & 94.85 \\
\stacktableoursrow
\textbf{Shared-budget allocation (ours)} & 98.97 & 97.51 & 95.26 \\
\bottomrule[0.9pt]
\end{tabular}
\caption{Static crop quotas versus StackTok's dynamic shared-budget allocation on LLaVA-NeXT-7B (average retained performance, \%).}
\label{tab:ablate-crop}
\end{table}

\noindent\textbf{Adjusted-threshold consequence.}
Because StackTok's calibrated $\beta$ lies in $[0,1]$, for any
$t\in\{1,\ldots,k\}$,
\begin{equation*}
C(\mathcal G_t)=C_t^{\mathrm{ref}}\ge\beta C_t^{\mathrm{ref}}=\theta_t.
\end{equation*}
Thus $\mathcal G_t$ witnesses feasibility of the size-$t$ support threshold. This
statement neither requires equality nor implies that StackTok's interleaved
prefix $S_t$ meets the same threshold.

\section{Additional Experimental Details and Results}
\label{sec:appendix-experiments}

\subsection{Evaluation Protocol}

\noindent\textbf{Affinity and calibration.} StackTok requires no training. Vision--vision affinity uses pre-projector features, while query--vision affinity uses post-projector embeddings. This separation measures coverage in the native vision space and relevance in the LLM input space. We set $\tau_t=0.02$ and $\tau_v=0.2$ across datasets, except that Qwen2.5-VL-7B uses $\tau_t=0.01$. Entropy evaluation uses the numerical floor $\delta=10^{-12}$, and the default strictness range is $[\beta_{\min},\beta_{\max}]=[0.3,0.9]$.

\noindent\textbf{Refinement and query processing.} Support-preserving swap refinement uses $\epsilon_{\mathrm{sw}}=10^{-6}$ and one pass for each local retained set containing at most 16 tokens. LLaVA query preprocessing retains content-bearing keywords before embedding, whereas the Qwen2.5-VL integration embeds the question-prefixed input text.

\subsection{Dynamic-Resolution Results}
Table~\ref{tab:qwen} tests whether reference-gated selection remains useful after Qwen2.5-VL-7B's native dynamic-resolution encoding. StackTok improves Avg.$^\dagger$ over MMTok by 1.21, 1.28, and 1.02 points at 20\%, 10\%, and 5\% retention, respectively, showing that the advantage persists when the visual sequence is not produced by a fixed grid. At 5\% retention, gains remain visible on MMB and POPE, whereas OCRBench is nearly tied (30.73 versus 30.70). This contrast suggests that the selector transfers across encoding regimes, while pointing to a remaining limitation in preserving fine text evidence after extreme compression.

\begin{table}[t]
\centering
\footnotesize
\stacktablestyle
\setlength{\tabcolsep}{5pt}
\begin{tabular}{lcc}
\toprule[0.9pt]
\stacktableheadrow
\textbf{Variant} & \textbf{8} & \textbf{16} \\
\midrule[0.45pt]
Without refinement & 84.8 & 90.4 \\
\stacktableoursrow
\textbf{Support-preserving swap refinement} & 85.2 & 90.7 \\
\bottomrule[0.9pt]
\end{tabular}
\caption{Effect of support-preserving swap refinement at two extreme-compression budgets on LLaVA-1.5-7B (average retained performance, \%).}
\label{tab:ablate-swap}
\end{table}

\subsection{Additional Ablations}
The following studies isolate choices beyond the component analysis in Table~\ref{tab:ablate-component}. All entries are average retained performance (\%).

\noindent\textbf{Reference-gated policy versus fixed scalarization.} Table~\ref{tab:ablate-constraint} compares StackTok with $R(S)+\alpha C(S)$, tuning $\alpha$ separately at each budget. StackTok still improves by 0.14, 0.10, and 0.06 points at 192, 128, and 64 tokens. Granting scalarization a separate best coefficient at each budget controls for budget-level tuning; the remaining gap indicates the limitation of a fixed coefficient: it cannot adjust the relevance--coverage priority across inputs and prefix states.

\noindent\textbf{Shared-budget allocation across crops.} Table~\ref{tab:ablate-crop} holds the crop-local selector fixed and changes only how the high-resolution budget is distributed. The gain over uniform quotas widens from 0.42 to 0.86 points as the budget contracts from 640 to 160 tokens, indicating that static allocation is most costly when every slot matters. Initial-$h_c$-proportional quotas recover part of this loss but remain 0.25--0.41 points behind. A one-time estimate may become stale as a crop accumulates tokens, whereas repeated nomination compares the current marginal value of the next token from each crop.

\noindent\textbf{Support-preserving swap refinement.} Table~\ref{tab:ablate-swap} compares the same reference-gated selector with and without the support-preserving swaps in Eq.~\eqref{eq:swap}. The 0.40- and 0.30-point gains at 8 and 16 tokens show that revisiting early greedy choices is useful when only a few selections can be retained. The gains are modest, consistent with refinement serving as a final correction while the coverage floor prevents that correction from sacrificing visual support.

\noindent\textbf{Strictness sensitivity.} Table~\ref{tab:hp} varies the clipping range $[\beta_{\min},\beta_{\max}]$ at the 64-token LLaVA-1.5-7B budget. The default $[0.3,0.9]$ performs best, while either neighboring range changes the result by at most 0.17 points. This narrow spread indicates local robustness rather than complete parameter insensitivity, since the sweep covers only one model--budget setting and nearby ranges.

\begin{table}[t]
\centering
\footnotesize
\stacktablestyle
\setlength{\tabcolsep}{3.5pt}
\begin{tabular}{lccc}
\toprule[0.9pt]
\stacktableheadrow
\textbf{$[\beta_{\min},\beta_{\max}]$} & $[0.2,0.8]$ & \cellcolor{stacksupport!11}\textbf{$[0.3,0.9]$} & $[0.4,1.0]$ \\
\midrule[0.45pt]
\textbf{Avg.\ \%} &  96.54 & \cellcolor{stacksupport!11}\textbf{96.66} & 96.49 \\
\bottomrule[0.9pt]
\end{tabular}
\caption{Sensitivity to the strictness clipping range on LLaVA-1.5-7B at 64 retained tokens (average retained performance, \%).}
\label{tab:hp}
\end{table}

\end{document}